\documentclass[letterpaper, 10 pt, conference]{ieeeconf}  % Comment this line out if you need a4paper

\IEEEoverridecommandlockouts                              % This command is only needed if 
\usepackage{graphics} % for pdf, bitmapped graphics files
\usepackage[tight,footnotesize]{subfigure} % for subfigures
\usepackage{epsfig} % for postscript graphics files
\usepackage{mathptmx} % assumes new font selection scheme installed
\usepackage{times} % assumes new font selection scheme installed
\usepackage{amsmath} % assumes amsmath package installed
\usepackage{amssymb}  % assumes amsmath package installed
\usepackage{balance} % balance references

\usepackage{glossaries}
\newacronym{ros}{ROS}{Robot Operating System}
\newacronym{mse}{MSE}{Mean Squared Error}
\newacronym{cnn}{CNN}{Convolutional Neural Network}
\newacronym{dqn}{DQN}{Deep Q-Network}
\newacronym{naf}{NAF}{Normalized Advantage Function}

\title{\LARGE \bf
Sensor Fusion for Robot Control through Deep Reinforcement Learning
}

\author{Steven Bohez, Tim Verbelen, Elias De Coninck, Bert Vankeirsbilck, Pieter Simoens and Bart Dhoedt $^{1}$% <-this % stops a space
\thanks{$^{1}$Authors are with Ghent University - imec, IDLab, Department of Information Technology.
        {\tt\small firstname.lastname@ugent.be}}%
}

\begin{document}

\maketitle
\thispagestyle{empty}
\pagestyle{empty}

%%%%%%%%%%%%%%%%%%%%%%%%%%%%%%%%%%%%%%%%%%%%%%%%%%%%%%%%%%%%%%%%%%%%%%%%%%%%%%%%
\begin{abstract}
Deep reinforcement learning is becoming increasingly popular for robot control algorithms, with the aim for a robot to self-learn useful feature representations from unstructured sensory input leading to the optimal actuation policy.
In addition to sensors mounted on the robot, sensors might also be deployed in the environment, although these might need to be accessed via an unreliable wireless connection.
In this paper, we demonstrate deep neural network architectures that are able to fuse information coming from multiple sensors and are robust to sensor failures at runtime.
We evaluate our method on a search and pick task for a robot both in simulation and the real world.
\end{abstract}

%%%%%%%%%%%%%%%%%%%%%%%%%%%%%%%%%%%%%%%%%%%%%%%%%%%%%%%%%%%%%%%%%%%%%%%%%%%%%%%%
\section{INTRODUCTION}

Deep reinforcement learning has recently been applied to transform raw sensory input to control signals for various robotic tasks such as locomotion~\cite{Schulman15}, grasping~\cite{Lenz15}, manipulation~\cite{Gu16}, autonomous driving~\cite{Hadsell08}, etc.
Generally the sensors and actuators are treated as a single, static system and the policy is trained on a fixed input size.
With the current trend towards the Internet-of-Things, more and more sensor inputs can be provided by the environment the (mobile) robot operates in.
However, off-board sensors are accessed over an unreliable wireless link, and it might not be known upfront which sensors are actually operational.

In this paper, we propose a deep learning approach to fuse information coming from multiple, possibly remote sensors, while being robust to sensors being unavailable at runtime.
A naive way of combining multiple inputs is by concatenating either the raw inputs or features calculated from these inputs in the neural network architecture~\cite{Ngiam11,Eitel15,Lenz15,Srivastava14}.
This approach however does not scale well in the number of inputs, and none define the behavior in the case of a missing input.
Our contribution is twofold.
First, we explore different neural network architectures for fusing multiple inputs that can be trained end-to-end with deep reinforcement learning methods.
Second, we use a DropPath regularization method during training to make the neural network robust to missing inputs at runtime.

We test our approach on a search and pick scenario, in which a robot consisting of an omnidirectional platform with a 5-DOF manipulator has to drive towards a can and pick it up.
As input, the robot receives lidar data from a sensor mounted on the robot, as well as sensors mounted in the environment.
Our neural networks are trained and evaluated in simulation, and we also present preliminary results transferring the trained policies to a real Kuka YouBot.

The remainder of the paper is structured as follows.
We discuss related work in Section~\ref{related}.
In Section~\ref{usecase} we present the detailed setup of our experiments.
Section~\ref{approach} presents our neural network architectures for fusing multiple inputs and DropPath regularization.
In Section~\ref{results} we present our experimental results and finally Section~\ref{conclusion} concludes this paper.

\section{RELATED WORK}
\label{related}

In this section, we will first discuss some important related work on both sensor fusion and deep reinforcement learning.

% put figure here so it is rendered on page 2
\begin{figure*}[t!]
\centering
\subfigure[]{
\includegraphics[height=2in]{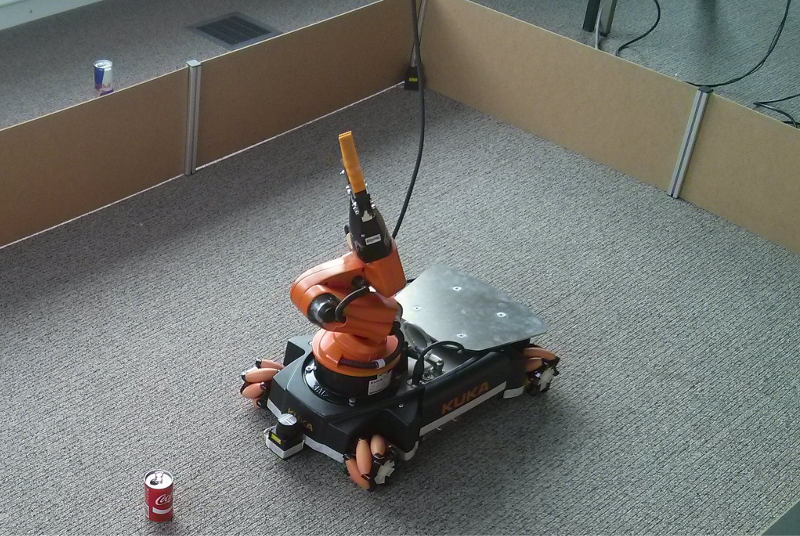}
\label{fig:real}
}
\subfigure[]{
\includegraphics[height=2in]{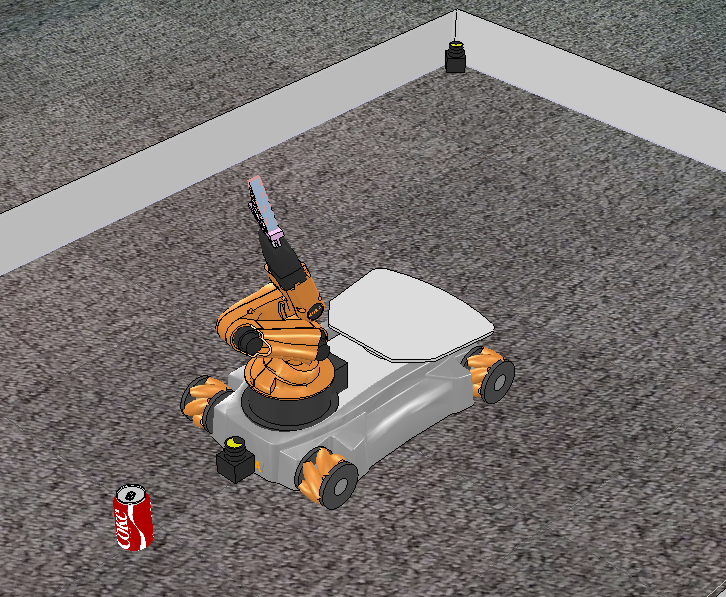}
\label{fig:sim}
}
\caption{Our real setup (a) with a Kuka Youbot in a rectangular cage, equipped with a Hokuyo lidar. Two more Hokuyo sensors are deployed in the corners of the cage forming the diagonal. We replicated the same setup in the V-REP simulator (b) for training.}
\label{fig:setup}
\end{figure*}

\subsection{Sensor Fusion}

Combining information from several sources in order to get a unified picture has been a research topic for decades~\cite{Khaleghi13}.
An often used sensor fusion technique is (Extended) Kalman Filtering and variants, which has been shown useful for use cases such as object position estimation~\cite{Stroupe01}, robot pose estimation~\cite{Dobrev16}, localization~\cite{Malyavej} and navigation~\cite{Lynen13}.
In these cases the desired state representation based on the sensor input (e.g. the robot pose) is known and fixed upfront.
However, in our approach we want to learn a policy end-to-end and the optimal feature representation to fuse the sensor input to is unknown upfront.

Work has been done on fusing multiple inputs using a deep learning approach, by learning a shared representation between all inputs~\cite{Ngiam11}.
Eitel et al.~\cite{Eitel15} present a two-stream \gls{cnn} for RGB-D object recognition.
Both RGB and D data are first separately processed by a pretrained \gls{cnn}, whose outputs are then concatenated and converged in a fully connected layer with softmax classifier.
In~\cite{Lenz15}, RGB-D data is used for detecting robotic grasps.
A structured regularization penalty is added when concatenating multi-modal features. Srivastava et al.~\cite{Srivastava14} fuse bi-modal image-text and audio-video data using deep Boltzman machines.
In this paper, we use neural networks to fuse lidar data from multiple sensors, which are trained end-to-end using a deep reinforcement learning approach. 

\subsection{Deep Reinforcement Learning}

Deep reinforcement learning uses deep neural network models as function approximators.
The neural network can either represent the policy itself by mapping observations to actions directly, or a value function from which a policy can be derived.
In this paper we mainly consider the action-value function $Q\left(s,a\right)$, defined as

\begin{equation}
Q\left(s,a\right) = \mathbb{E}_{\pi}\left[\sum_{k=t}^{T} \gamma^{k-t} r_{k} \middle| s_t=s,a_t=a \right].
\end{equation}

$Q\left(s,a\right)$ represents the expected discounted return, or $Q$-value, of executing an action $a$ in a given state $s$ and executing policy $\pi$ afterwards.
The reward $r$ in each time step is discounted with a factor $\gamma \in \left[0,1\right]$, which gives more weight to short-term rewards.
The policy $\pi\left(a\middle|s\right)$ determines which action to take in each state.
The optimal policy is found when following the optimal $Q$ function $Q^* = \max_{\pi} \mathbb{E}_{\pi}\left[Q(s,a)\right]$.
In Q-learning, $Q^*$ is typically found by iteratively taking the action with the highest $Q$-value, i.e. $\pi\left(a\middle|s\right) = {\arg\max}_{a} Q\left(s,a\right)$, until convergence.

\glspl{dqn} extend Q-learning by approximating the Q-function by a deep neural network and have achieved state-of-the-art results in playing ATARI games~\cite{Mnih15}.
This method uses the game pixel data as input and outputs discrete actions.
The performance of this method was further improved using double Q-learning~\cite{Hasselt10} and prioritized experience replay~\cite{Schaul15}.

Recently, deep Q-learning was extended for continuous actions using \glspl{naf}~\cite{Gu16}.
Other methods that work on continuous action spaces have an explicit policy function and directly optimize the expected reward using the policy gradient~\cite{Schulman15} or use an actor-critic approach in which a separate actor and critic neural network are used~\cite{Lillicrap15}.
The actor network converts input to actions, while the critic network evaluates the actor network and provides gradient information.
This approach can also be extended to stochastic policies~\cite{Heess15}.

In the remainder of this paper we will use the \gls{dqn} algorithm for training our policies.
However, our neural network architecture for fusing multiple inputs might equally well be applied to the other deep reinforcement learning algorithms.

\section{ENVIRONMENT SETUP}
\label{usecase}

As a test case for our sensor fusion approach, we have created a controlled environment for a search and pick task.
Our setup is depicted in Figure~\ref{fig:real}.
We created a 1.8 by 2.7 meter rectangular arena in which a Kuka YouBot is placed.
The robot consists of an omni-directional base and a 5-DOF manipulator.
A two-finger soft gripper is mounted on the manipulator to facilitate object picking.
As a pick target we use soda cans.
Sensor information is provided by Hokuyo URG-04LX-UG01 1D lidar sensors, of which one is mounted on the robot base, and two more are mounted on nonadjacent the arena corners.
All sensors are configured to have a 180 degree field of view.
An Nvidia Jetson TX1 embedded GPU is mounted inside the YouBot base and provides the compute power required for evaluating the policy.
Each environment lidar is attached to a Raspberry Pi 2 that is connected over WiFi to the Jetson board.
Both robot and sensors are controlled using the \gls{ros}~\cite{ros}.

We have also replicated this environment in simulation as shown on Figure~\ref{fig:sim}, using the V-REP simulator~\cite{vrep}.
Within the simulator, the robot and sensors expose exactly the same interface via ROS.
This allows us to control the simulated environment in the same way as the real world.

\section{NEURAL NETWORK ARCHITECTURES FOR DEEP RL SENSOR FUSION}
\label{approach}

\begin{figure*}[t!]
\centering
\includegraphics[height=1.8in]{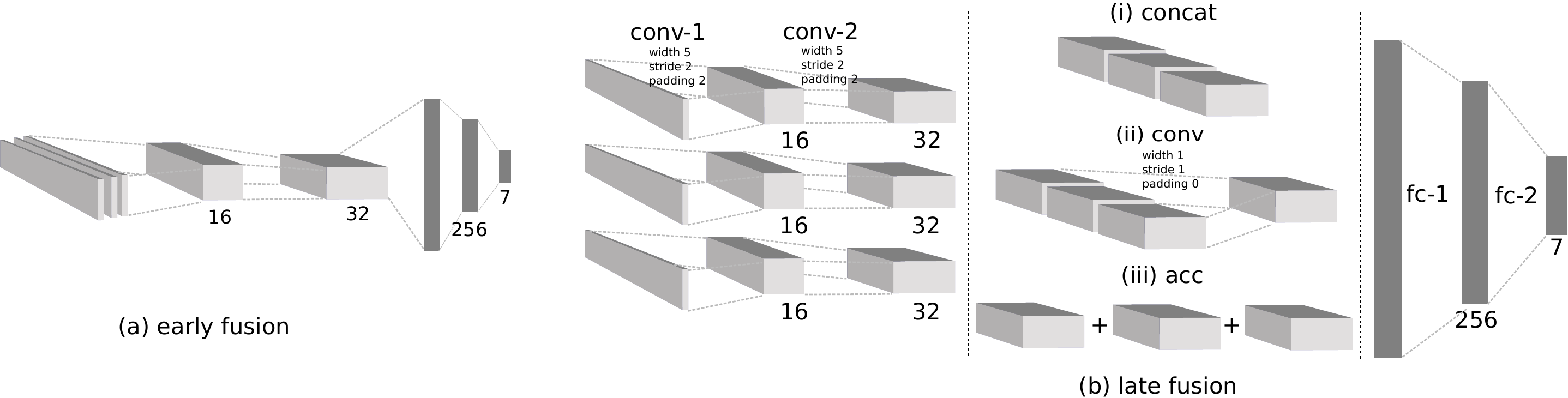}
\caption{The early fusion architectures (a) concat all inputs as feature planes. The late fusion architectures (b) start with two convolution layers for each sensor input, which is then merged for the fully connected layers. We present three types of late fusion: (i) simply concatenating the convolution outputs, (ii) reducing the concatenated feature planes with a 1x1 convolution, and (iii) accumulating the convolution outputs.}
\label{fig:fuse}
\end{figure*}

We will use a \gls{dqn}~\cite{Mnih15} to approximate and maximize the action-value function Q and derive the policy.

In our case the state $s$ is determined by the observations of the three lidar sensors.
Hence, our \gls{dqn} will consist of three inputs, one for each sensor, and of one output: the $Q$ values for each of the actions.
For simplicity we have discretized the action space to seven actions for the robot: move left/right, move forward/backward, rotate clockwise/counterclockwise and execute a (hard-coded) grip.
Not only do we want to fuse the input coming from the three sensors, we also want to be robust against sensor failures, since two of the sensors are not wired to the robot.
However, we want to avoid unnecessarily increasing the number of weights when new sensors are added.

There are multiple ways to combine multiple inputs in a neural network.
As also pointed out by~\cite{Lenz15}, inputs can be concatenated to a single input for a neural network.
This approach will be denoted as ``early fusion''.
A second approach is to first process the inputs in separate neural network stacks, that are later combined into a number of common layers, which we will call ``late fusion''.

\subsection{Early fusion}

We start by defining a baseline model, that uses only the sensor mounted on the robot as input.
Similar to~\cite{Mnih15}, this model starts with two convolution layers, the first one with 16 filters, the second one with 32 filters, followed by a fully connected layer with 256 hidden units and a final fully connected layer that produces 7 outputs.
In each layer we use ReLU activations. 

Next, we add the two additional sensor inputs by early fusing.
A first, small model is similar to single-sensor model, but now the additional sensor inputs are added as additional feature planes to the first convolution.
A second, large model triples the number of filters and hidden units in the fully connected layer to compensate for the additional inputs.

\subsection{Late fusion}

In our late fusion models we dedicate the first two convolution layers to each sensor input, and fuse the outcoming feature representations before forwarding to the fully connected layers, as shown in Figure~\ref{fig:fuse}.
In this case, the ReLU activation of the second convolution layer is applied after the merge, before the fully connected layer.
A naive approach of fusing is to concatenate the convolution outputs into a single feature vector.
This has the main disadvantage of dramatically increasing the number of weights of the next fully-connected layer as the number of sensors increases.
This is because we fuse at the layer with the highest number of activations.
In order to overcome this effect somewhat, we introduce as second model that adds an additional convolutional layer with width 1 to first collapse the concatenated feature planes.
Finally, we propose a model that accumulates the convolution outputs directly, which requires no additional parameters but also limits the connectivity between the feature maps from different sensors.
The intuition is that by training this model end-to-end, the output of the second convolution layer of each sensor will be a suitable state representation where each sensor can add weight to each of the features.
Features that get enough weight from one or more sensor inputs will be activated for the fully connected layer.

An overview of all models and the number of parameters is listed in Table~\ref{tab:models}.

\begin{table}[b!]
\centering
% increase table row spacing, adjust to taste
\renewcommand{\arraystretch}{1.3}
\caption{Overview of the different policy architectures}
\label{tab:models}
\begin{tabular}{|c|c|c|}
\hline
\textbf{Model} & \textbf{Architecture}  & \textbf{\#Params}  \\ 
\hline
\textit{Single} & conv(16)\textgreater conv(32)\textgreater fc(256)\textgreater fc(7) & 266887 \\
\hline
\hline
\textit{Early (small)} & conv(16)\textgreater conv(32)\textgreater fc(256)\textgreater fc(7) & 267047 \\
\hline
\textit{Early (large)} & conv(48)\textgreater conv(96)\textgreater fc(256)\textgreater fc(7) & 812391 \\
\hline
\hline
\textit{Late (concat)} & concat after 3x conv(32) & 796551 \\
\hline
\textit{Late (conv)} & conv (width 1) (32) after concat & 275367 \\
\hline
\textit{Late (acc)} & accumulate after 3x conv(32) & 272263 \\
\hline 
\textit{Acc + DP} & add DropPath before accumulate & 272263 \\
\hline
\end{tabular}
\end{table}

\subsection{DQN training and DropPath regularization}

To train each model, we let agents interact with the environment and store each tuple ($s$,$a$,$r$,$s'$) in an experience pool $\mathcal{P}$.
The Q-network is then updated by sampling from the experience pool and minimizing the loss

\begin{equation}
L_{i}\left(\theta_i\right) = \mathbb{E}_{\left(s,a,r,s'\right)\sim\mathcal{P}}\left[r + \gamma \max_{a'}Q\left(s',a';\bar{\theta_{i}}\right) - Q\left(s,a;\theta_i\right) \right]^2,
\label{bellmann}
\end{equation}

where $\theta_i$ are the Q-network parameters and $\bar{\theta_i}$ are the parameters of the slow moving target as in~\cite{Mnih15}.

In our environment, chances are that one or more sensor inputs are unavailable at a given moment in time.
In order to have a policy that is robust against unavailable sensor information, we introduce DropPath regularization, similar to what is used in training FractalNets~\cite{Larsson16}.
We apply DropPath in our \textit{Late (acc)} model, by dropping each of the environment sensors input with a certain probability before the accumulation. 

However, applying DropPath during the policy training with Equation~\ref{bellmann} did not yield good results.
Instead, we first train a Q-network until convergence using Equation~\ref{bellmann}, resulting in trained parameters $\theta^{*}$.
Next, we add DropPath, and further train the Q-network by minimizing the loss

\begin{equation}
L_{j}\left(\theta_j\right) = \mathbb{E}_{\left(s,a,r,s'\right)\sim\mathcal{P}}\left[ Q\left(s,a;\theta^{*}\right) - Q\left(s,a;\theta_j\right) \right]^2
\label{refine}
\end{equation}

where $Q\left(s,a;\theta^{*}\right)$ is the output of the original converged Q-network with all sensor information available.
So intuitively, we first approximate the optimal $Q$ function using \gls{dqn}, and next we try to reconstruct this $Q$ function with missing sensor input.

\section{EXPERIMENTAL RESULTS}
\label{results}

To speed up the learning process, we simulate multiple agents that are uploading concurrently to a single, shared experience pool, and limit the resolution of the simulated lidar sensors to 128 scan points.
The simulator steps at 100ms, which is the update rate of the lidar sensors, and each simulation step is recorded as experience pool sample. 
We run simulations on a compute cluster of 40 virtual machines with 2 CPUs and 4GB RAM each.
We don't require GPU infrastructure for training since the limiting factor is the experience generation using the V-REP simulator.

For each simulation, a random start position and orientation for both the robot and soda can are generated.
An optimal target location for the soda can is right in front of the robot, where a hard-coded grip action is able to pick up the can.
The goal then becomes for the robot to position itself that the can is on the optimal target spot. 

As the reward function, we use the normalized negative distance of the can position to the optimal target location of the robot.
This results in a reward of -1 when the robot is located at one end of the arena and the can is at the opposite end, and a reward of 0 when the robot can actually fetch the can with the hard coded grip action.
A fixed reward of -1 is given when the robot collides with either the border or the can.
Every roll-out is capped at 100 simulation steps, which is enough for the robot to fetch the can from any start configuration.

\begin{figure}[t!]
\centering
\includegraphics[width=3.5in]{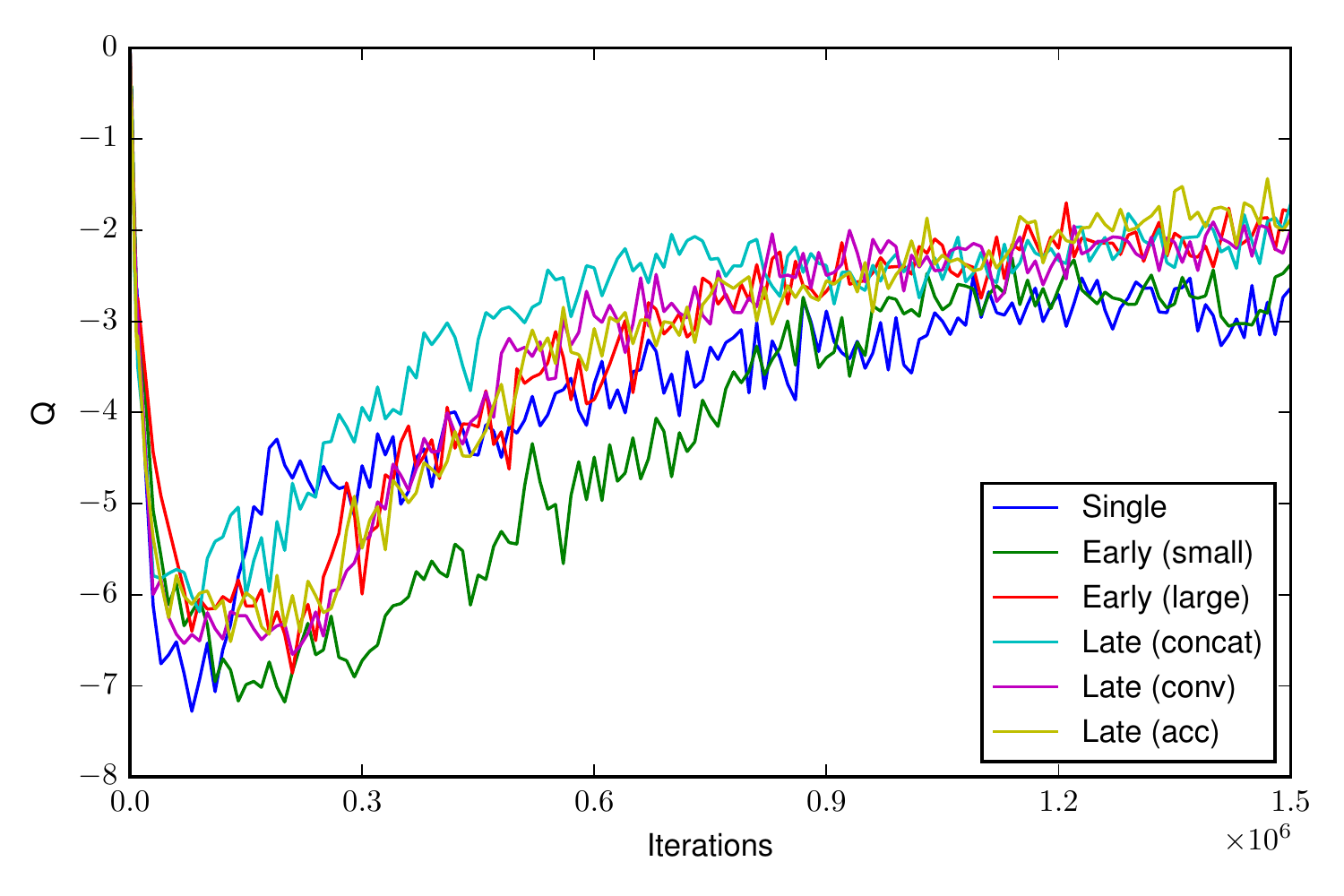}
\caption{Average $max_a Q\left(s,a\right)$ of mini-batches during training. After 1.5M iterations the Q-networks converge. Also note that the \textit{Single} and \textit{Early (small)} converge to a slightly lower $Q$-value.}
\label{fig:results3}
\end{figure}

All models are trained with DIANNE~\cite{DeConinck15}, a distributed deep learning framework built on Torch~\cite{torch}, that has built-in support for deep reinforcement learning.
We update the neural network weights according to Equation~\ref{bellmann}, but we use double Q learning as described in~\cite{Hasselt10}, and pseudo-Huber loss instead of \gls{mse} as it is more resilient to outliers.
For the gradient updates we use RMSProp with a learning rate of 0.0001.
All agents use $\epsilon$-greedy exploration with $\epsilon=1$ at the start and exponentially annealed to $\epsilon=0.1$.
Our experience pool has a maximum size of 1 million samples, and is updated in a first-in-first-out manner.
We train our models for 1.5 million batches of size 32. The average $max_a Q\left(s,a\right)$ per mini-batch is plotted on Figure~\ref{fig:results3}. Note that the \textit{Single} and \textit{Early (small)} converge to a slightly lower $Q$-value, hinting at a lower success rate.

\begin{figure}[b!]
\centering
\includegraphics[width=3.5in]{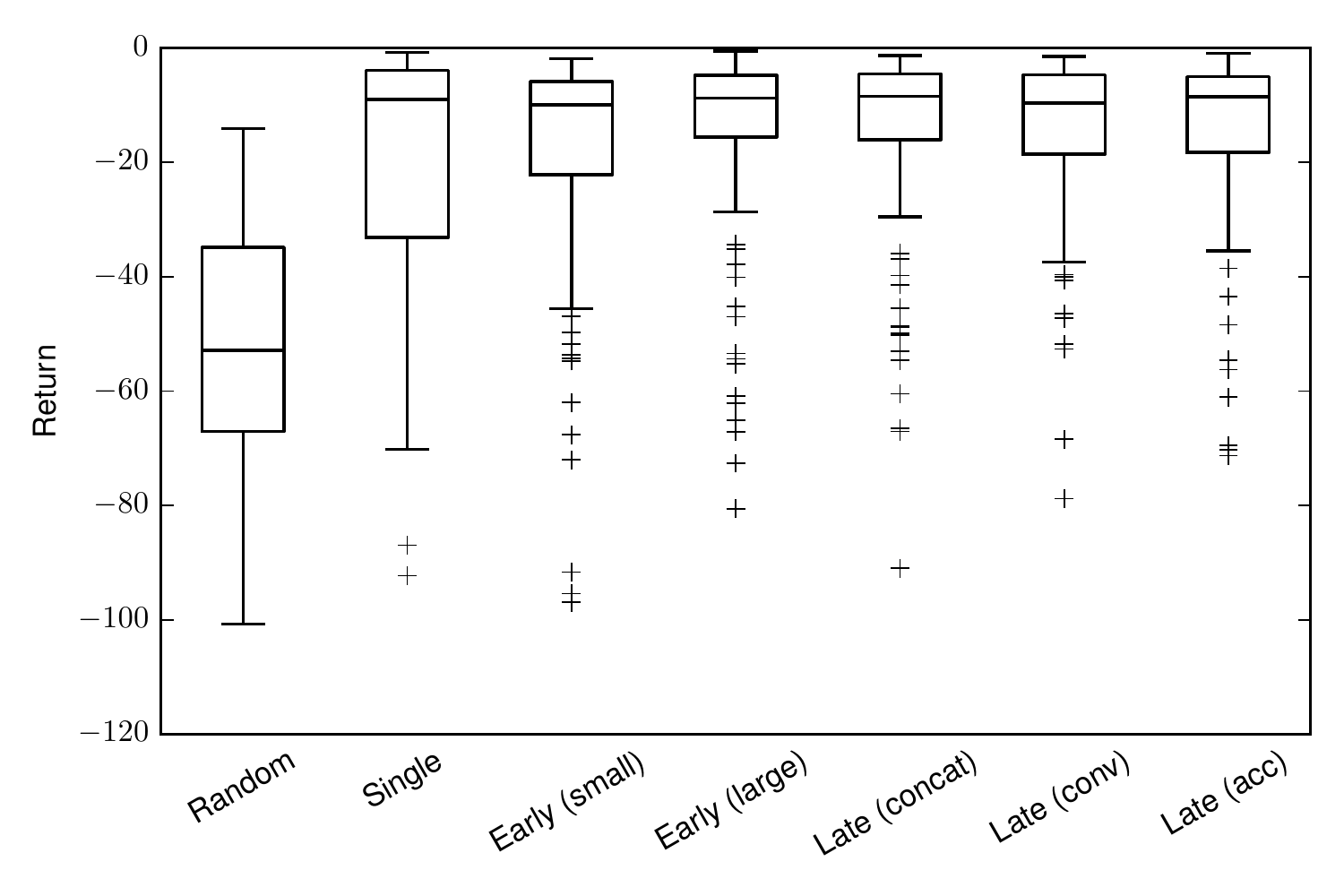}
\caption{Box plot of the returns achieved on 100 roll-outs for each of the models. The Q-network clearly benefits from additional sensor information, although there is little difference between late or early fusion. The models with more parameters perform slightly better.}
\label{fig:results1}
\end{figure}

\subsection{Early vs late fusion}

To compare whether our models are able to fuse information coming from multiple sensors, we compare them against two baselines: one taking purely random actions, and the \textit{Single} model that is trained on the robot's front-facing sensor only.
To evaluate the models' effectiveness, we randomly generated 100 start poses for the robot and the can, evaluated each policy for each of these start configurations and capture the total return collected by the agent in each roll-out.
Since the absolute reward depends on the start situation, we compare box plots showing the distribution of the rewards per roll-out.

Our results are shown in Figure~\ref{fig:results1}.
It is clear that the \textit{Single} model performs much better than the pure random baseline.
Adding information from the two additional lidar sensors introduces an additional leap in performance, showing that the models are indeed able to fuse the information.
More specific, we see that in starting situations where the can is positioned behind the YouBot (out of view of the front-facing sensor) the \textit{Single} model is unable to correctly navigate towards the can, whereas this is more often the case in the fused models.
We observed no significant difference in performance between early or late fusion models.
The two models with a larger amount of parameters in the first linear layer (\textit{Early (large)} and \textit{Late (concat)}) perform slightly better than the others.
Also note the large negative tails of all box plots, indicating that there are certain start configurations that none of the models are able to solve correctly.

\subsection{DropPath regularization}

To evaluate DropPath regularization, we add DropPath to our \textit{Late (acc)} model, with a drop rate of 0.5 for the remote sensors.
The front facing sensor input is never dropped, as we assume this one is always available since it is directly wired into the robot.
This balances equally between only having the front facing sensor, front facing + sensor 1, front facing + sensor 2 and all sensors available.
We refine the \textit{Late (acc)} model with additional training using Equation~\ref{refine} for another 1M mini-batches on 5M experience pool samples generated with \textit{Late (acc)}, resulting in the \textit{Acc + DP} model.

\begin{figure}[b!]
\centering
\includegraphics[width=3.5in]{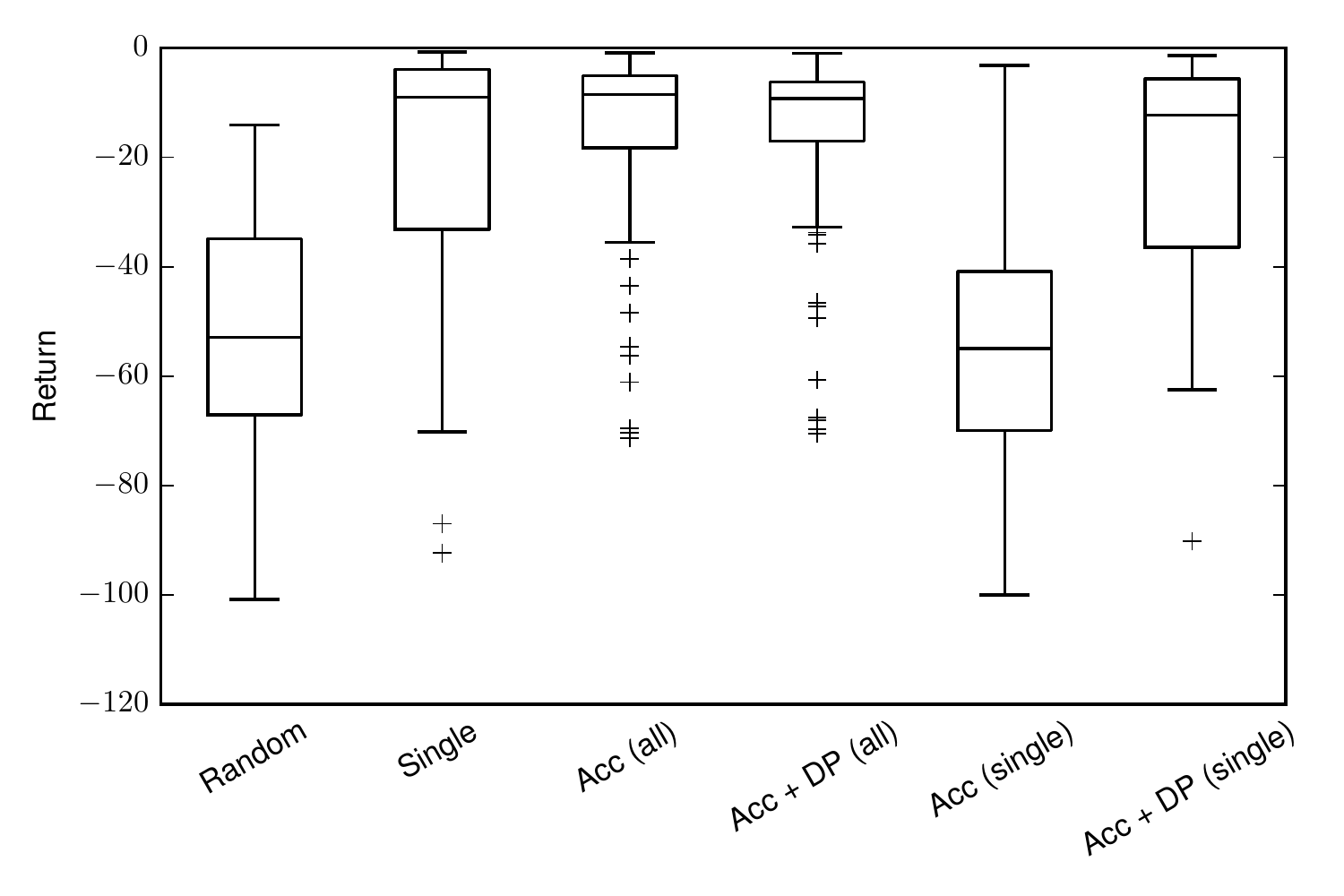}
\caption{When removing the remote sensor information from \textit{Late (acc)}, this performs worse than random. However, when refining with DropPath regularization, we get similar results as training on single sensor data, while still performing better when all sensors are present.}
\label{fig:results2}
\end{figure}

We compare the regularized \textit{Acc + DP} model to \textit{Late (acc)} when using all three or only the front facing sensor in Figure~\ref{fig:results2}.
Since \textit{Late (acc)} was trained on input data from all three sensors, we see that dropping the environment sensors has serious impact on the performance.
When dropping the environment sensors, \textit{Late (acc)} performs even worse than random.
After regularization with DropPath however, we see that \textit{Acc + DP} performs on par with \textit{Single} when only the front facing sensor is available.
When all sensors are available, there is only a negligible performance difference with the original \textit{Late (acc)} model.
This shows that our \textit{Acc + DP} model is indeed able to fuse information from multiple sensor sources, while still being robust to missing sensor information at runtime.

\begin{figure}
\centering
\includegraphics[width=3.5in]{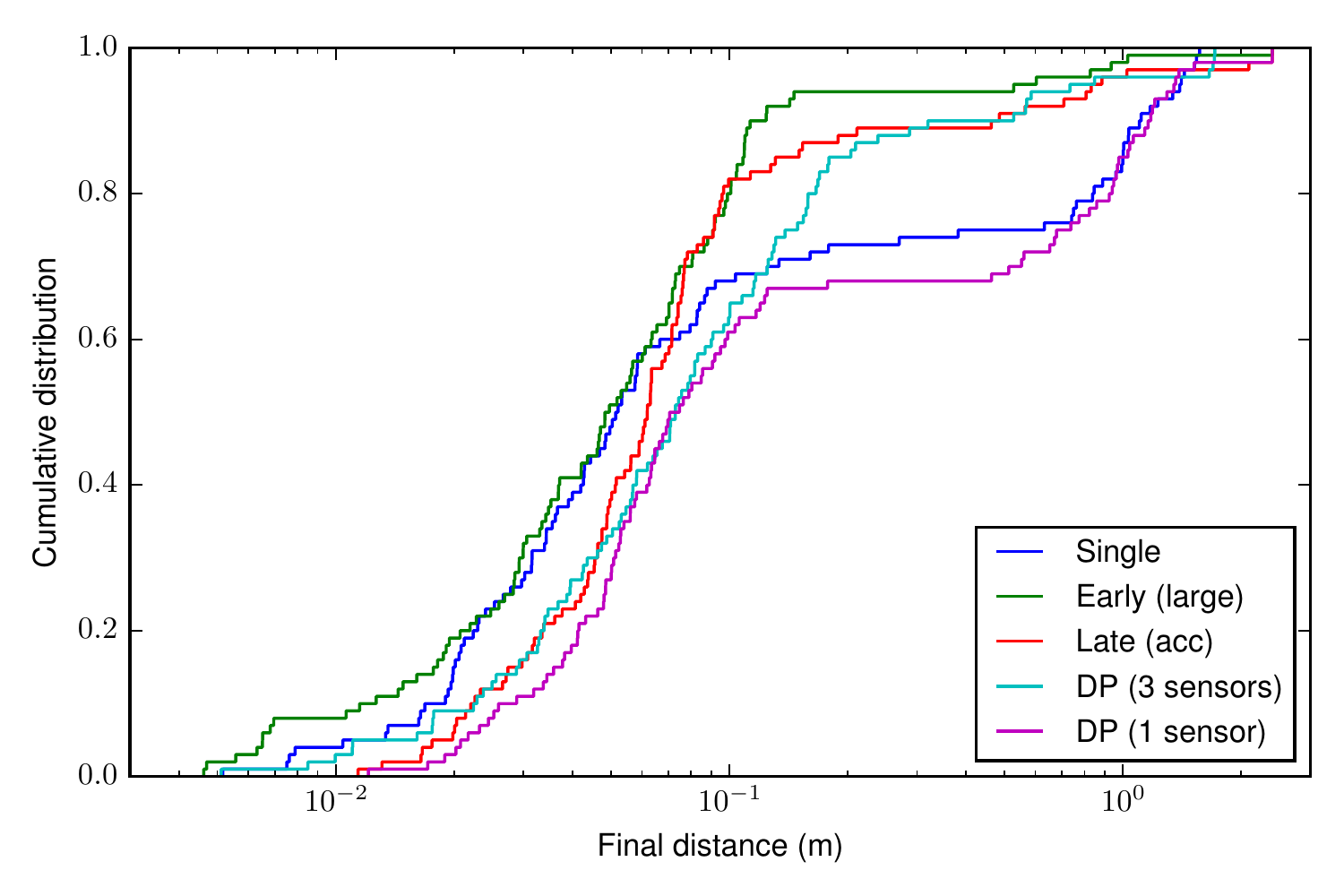}
\caption{Cumulative distribution of the final distance of the can to the optimal grip location.}
\label{fig:results4}
\end{figure}

To evaluate the final position of the robot, we have plotted on Figure~\ref{fig:results4} the cumulative distribution of the final distance of the can to the ``optimal'' grip location where a reward of 0 is received. All methods end in a 10 cm radius of this location in more than half of the cases. The models that only have a single sensor input perform worse than the methods fusing multiple sensors, often missing out in the cases that the can is positioned behind the robot.

\subsection{Real-world results}

When transferring our trained models to our real-world setup, we experienced a number of problems.
First, with a fixed grip action it is difficult for the robot to actually pick up a can with the two-finger gripper, as the robot has to be positioned very accurately.
Second, the lidars' rays don't reflect on all material, and reflection also depends on the angle.
Especially the YouBot's base seemed not to reflect the lidar beams well, resulting in missing values in the sensor inputs (which are set to 0).

We mitigated the first issue by adapting the grip action to take feedback from the last lidar scan to estimate the exact position of the can to the robot base.
To make the YouBot more ``visible'' by the environment sensors, we covered the base with masking tape.
However, imperfect scans with missing measurements still occurred.
Therefore, we also added DropOut~\cite{srivastava14a} on the sensor inputs during our DropPath refinement, dropping lidar rays from the input with a probability of 0.025.
This made our \textit{Acc + DP} model more robust against the real-world lidar data.
Results of our real-world experiments can be seen in the supplementary video material.

\section{CONCLUSION}
\label{conclusion}

In this paper, we presented methods for fusing sensor input from multiple sources into a robot control policy using deep reinforcement learning.
We evaluated different architectures on a search and pick task with a Kuka YouBot and multiple lidar sensors.
We have shown that neural networks are indeed able to fuse information from multiple sources.
We have found no significant performance difference between fusing early or late in the neural network.
However, combining our late fusion approach by means of accumulation with DropPath regularization, we are able to train a policy network that leverages all sensor inputs, while still being robust when sensor input is missing.

As future work we will extend our evaluation to a wide variety of deep reinforcement learning algorithms, more specifically exploring recurrent policies that can better capture the environment dynamics, as well as using methods that work on continuous action spaces as this would be more suited for controlling the robot.
Next, we would also like to expand our approach to multi-modal sensor input, for example combining the lidar data with RGB(-D) camera input.

%%%%%%%%%%%%%%%%%%%%%%%%%%%%%%%%%%%%%%%%%%%%%%%%%%%%%%%%%%%%%%%%%%%%%%%%%%%%%%%%

%%%%%%%%%%%%%%%%%%%%%%%%%%%%%%%%%%%%%%%%%%%%%%%%%%%%%%%%%%%%%%%%%%%%%%%%%%%%%%%%

%%%%%%%%%%%%%%%%%%%%%%%%%%%%%%%%%%%%%%%%%%%%%%%%%%%%%%%%%%%%%%%%%%%%%%%%%%%%%%%%

%\section*{APPENDIX}
%
%Appendixes should appear before the acknowledgme
\section*{ACKNOWLEDGMENT}

Steven Bohez is funded by Ph.D. grant of the Agency of Innovation by Science and Technology in Flanders (IWT).
We gratefully acknowledge the support of NVIDIA Corporation with the donation of the Jetson TX1 used in this work.

% balance columns
\balance

%%%%%%%%%%%%%%%%%%%%%%%%%%%%%%%%%%%%%%%%%%%%%%%%%%%%%%%%%%%%%%%%%%%%%%%%%%%%%%%%

\bibliographystyle{plain}
\bibliography{bibtex}

\end{document}